\documentclass[a4paper]{article}

\usepackage{INTERSPEECH2021}

\title{LRWR: Large-Scale Benchmark for Lip Reading in Russian language}
\name{Evgeniy Egorov$^{1*}$\thanks{*Work done during an internship at Tinkoff. \newline\newline Preprint. Under review.}, Vasily Kostyumov$^{1*}$, Mikhail Konyk$^{1*}$, Sergey Kolesnikov$^2$}
\address{
  $^1$Moscow Institute of Physics and Technology\\
  $^2$Tinkoff.AI
}
\email{egorov.ev@phystech.edu, kostyumov.vv@phystech.edu, konyk.ma@phystech.edu, s.s.kolesnikov@tinkoff.ai}

\begin{document}

\maketitle

%
\begin{abstract}

Lipreading, also known as visual speech recognition, aims to identify the speech content from videos by analyzing the visual deformations of lips and nearby areas.
One of the significant obstacles for research in this field is the lack of proper datasets for a wide variety of languages: so far, these methods have been focused only on English or Chinese.
In this paper, we introduce a naturally distributed large-scale benchmark for lipreading in Russian language, named LRWR, which contains 235 classes and 135 speakers. We provide a detailed description of the dataset collection pipeline and dataset statistics.
We also present a comprehensive comparison of the current popular lipreading methods on LRWR and conduct a detailed analysis of their performance. 
The results demonstrate the differences between the benchmarked languages and provide several promising directions for lipreading models finetuning. Thanks to our findings, we also achieved new state-of-the-art results on the LRW benchmark.

\end{abstract}

\section{Introduction}

In recent years, the machine learning community has made significant progress in solving the problem of speech recognition from audio.
Deep learning methods for Automatic Speech Recognition are now showing impressive performance but often struggle to understand the talk in noisy environments or the case of multiple speakers. 
Adding lipreading methods can help in solving this problem if there is a video  with a speaker. 
Nowadays, when lot's of people use voice messages while walking or driving, news reports are filmed directly from the place of events, such an expansion of speech recognition systems is becoming more and more valuable.
The word-level approach to solve the Lipreading task is recognizing individual words from a previously known dictionary. Still, even with this limitation, visual recognition is a challenging task for computer vision. However, with the advent of deep learning, there has been significant progress in this task.

Besides all the above progress, the primary catalyst for developing new lipreading methods and architectures that could qualitatively solve the speech recognition problem was the emergence of large datasets collected from non-laboratory data.
The first such dataset is LRW  \cite{Chung16}, proposed in 2016, which consists of 500 classes and is based on BBC news footage. Another available benchmark for lipreading in Chinese is the LRW-1000 \cite{DBLP:journals/corr/abs-1810-06990}, proposed in 2018, which consists of 1000 classes and displays a wide variety of speech conditions. 
Unfortunately, there was no such dataset for Russian language.

\begin{table*}[htb]
  \caption{Datasets statistics}
  \label{tab:datasets}
  \centering
  \begin{tabular}{lcccr}
    \toprule
    \multicolumn{1}{c}{\textbf{Datasets}} & 
    \multicolumn{1}{c}{\textbf{number of classes}} &
    \multicolumn{1}{c}{\textbf{number of speakers}} &
    \multicolumn{1}{c}{\textbf{Resolution}} &
    \multicolumn{1}{c}{\textbf{Environment}} \\
    \midrule
    OuluVS2: \cite{article}   & 10 & 53 & Fixed, with 6 different sizes & Lab. ~~~             \\
    LRW: \cite{Chung16}   & 500  & \textgreater1000 & Fixed ($256\times256$) & TV ~~~            \\
    LRW-1000: \cite{DBLP:journals/corr/abs-1810-06990}   & 1000 & \textgreater2000 & Naturally distributed & TV ~~~            \\
    LRWR (ours)   & 235 & 135 & Fixed ($112\times112$) & YouTube ~~~       \\
    \bottomrule
  \end{tabular}
\end{table*}

To this end, we collect a naturally-distributed large-scale dataset for lipreading in Russian language and provide a comprehensive comparison of the current popular lipreading methods. The contributions in this paper are summarized as follows.

\begin{figure}[t]
  \centering
  \includegraphics[width=\linewidth]{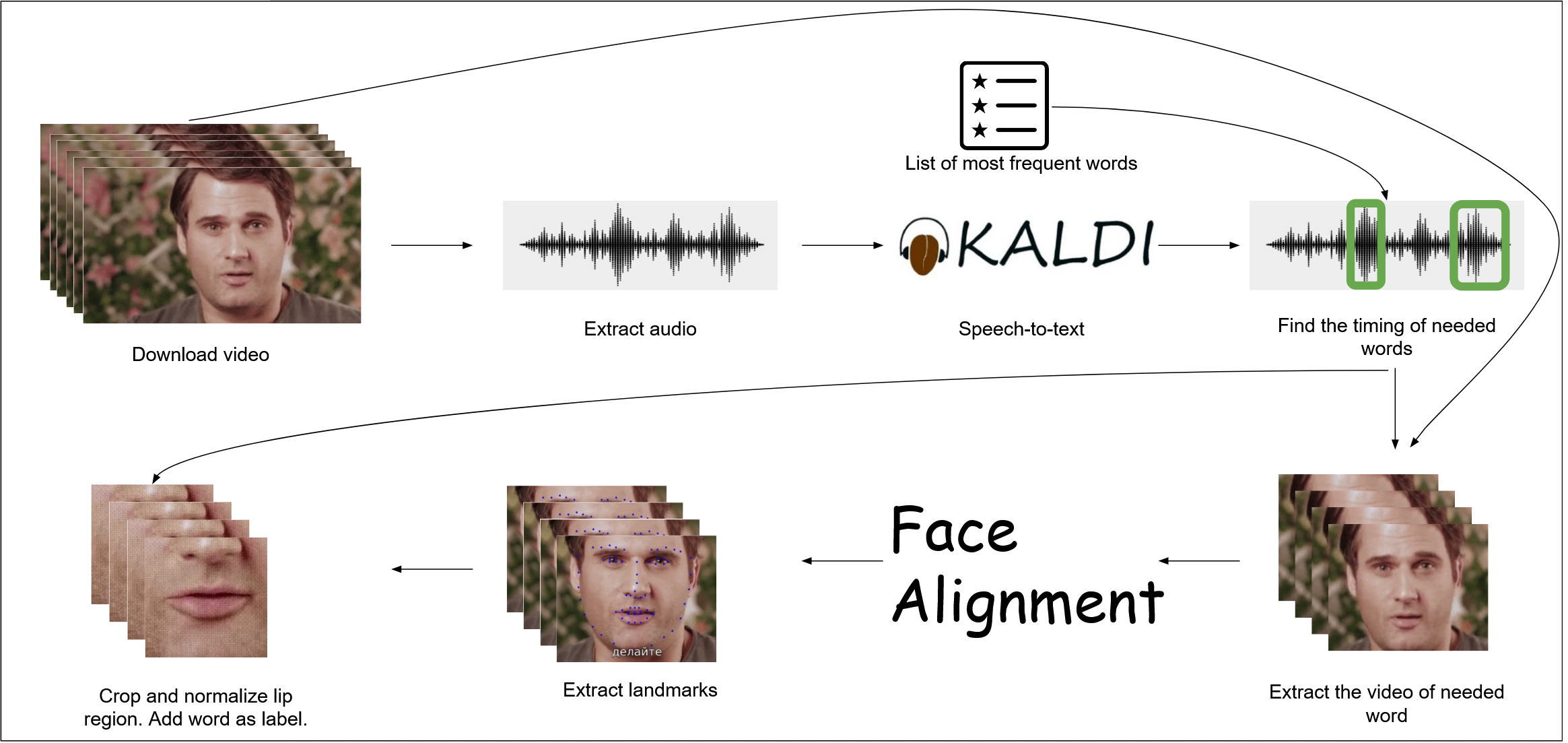}
  \caption{Data collection pipeline}
  \label{fig:speech_production}
\end{figure}

Our first contribution is a challenging and naturally distributed dataset for lipreading in Russian language – LRWR. (section 3).
There are 235 words from 135 speakers, with over 117500 samples in total. 
To the best of our knowledge, it's the first publicly available Russian lipreading dataset and Slavic-based language dataset.

Our second contribution is benchmarking a various number of existing architectures that demonstrate the best results on LRW and LRW-1000, such as deformation Flow Based Two-Stream Network \cite{xiao2020deformation}, D3D \cite{DBLP:journals/corr/abs-1810-06990} and other architectures \cite{martinez2020lipreading, feng2020learn} on LRWR (section 4). Besides, we conduct a comprehensive quantitative study and comparative analysis of the impact of different deep learning techniques on the resulting lipreading methods performance (section 4.3). In our analysis, we review different regularization and augmentation techniques, as well as various methods for data sampling and learning rate scheduling during model training. 

The final point of our study is the proposal of a new architecture that summarises our experiments' findings and achieves the best results on our dataset and the open LRW dataset, improving the previous state-of-the-art result from 88.4\% \cite{feng2020learn} to 89.1\% accuracy. (section 4.4) 

We hope that the open-sourced LRWR dataset and the proposed architecture will serve as the foundation for developing lipreading techniques in Russian and other languages of Commonwealth of Independent States.

\section{Related work}
Large-scale datasets have successively proven their fundamental importance in many research fields. Word-level lipreading was no exception, and most recent breakthroughs came with the emergence of such datasets as the LRW or the LRW-1000.
In this section, we observe the popular datasets and methods being state-of-the-art or close to them based on results obtained on the datasets for word-level lipreading.

\subsection{Datasets}
Several most popular world-level lipreading datasets are summarized in Table \ref{tab:datasets}. All of them have a significant impact on the deep learning progress in lipreading. In this part, we will give a short overview of these datasets shown in the table.

OuluVS2 \cite{article}, released in 2015, consists of 10 phrases spoken by 53 speakers with 9109 samples in total. This dataset was filmed in a laboratory environment and contained various camera angles: frontal, profile, 30\textdegree, 45\textdegree, 60\textdegree.
However, due to laboratory conditions, there are only a few variations beyond the view positions, and the viewpoints are always identical, which leads to poor robustness under real-life conditions for OuluVS2-trained networks.

LRW: \cite{Chung16}, the first large-scale dataset, released in 2016, contains 500 classes with more than 1000 recorded pronunciations per class. 
This dataset no longer consists of videos with word-by-word pronunciation recorded in a controlled lab environment but utilizes data from the BBC channel, which gives various speakers, angles, and lighting conditions. 
All this together makes it possible to use the LRW-trained network in the real world. 

LRW-1000 \cite{DBLP:journals/corr/abs-1810-06990}, the largest dataset available for lipreading, released in 2018 and contains 1000 classes and more than 2000 individual speakers. 
The dataset collection pipeline was optimized for real-world applications. Thus it aims at covering a "natural" variability over different speech modes and imaging conditions.

Although there have been many lipreading datasets as listed above, there are very few lipreading datasets for Slavic languages available up to now. 
Therefore, we hope LRWR could help future ASR research in Russian.

\subsection{Methods}

Over the last decade, there was a lot of progress in lipreading methodology. Early methods solved the Lipreading problem with expertly derived techniques using mathematical image transformations such as Discrete Cosine Transform (DCT) \cite{inproceedings_CDT}, Active Appearance Model (AAM) \cite{sterpu2018lipreading_AAM}
and Local Binary Pattern (LBP) \cite{article_LBP}.

With the advent of the LRW dataset, it became possible to train deep neural networks and utilize them for real-world applications effectively.
At first, a lot of research was focused on 2D fully convolutional networks \cite{article_only2dconv}. Nevertheless, thanks to hardware evolution, it soon became possible to use 3D convolution over 2D convolutions \cite{Chung16} or recurrent neural networks \cite{DBLP:journals/corr/StafylakisT17} for more efficient temporal information usage.
This idea has grown into a type of architecture where information about the local lip movement is extracted using 3D+2D convolutions backbone, while final temporal information is summarized using recurrent layers at the end.
This approach has been extremely successful, and many state-of-the-art lipreading solutions used it as a meta-architecture even nowadays \cite{feng2020learn}.

In addition to the above methods, there have been many attempts to modify architecture:
\begin{itemize}
\item D3D architecture \cite{DBLP:journals/corr/abs-1810-06990}, where 3D+2D convolutions are replaced by 3D only. 
\item Two-branch network \cite{xiao2020deformation}, which takes into account the temporal component using Deformaition-Flow, thus replacing the initial 3D convolutions in one branch.
\end{itemize}
Moreover, the researchers experimented with the replacement of recurrent layers: temporal convolutional layers \cite{martinez2020lipreading}, and Transformer \cite{wiriyathammabhum2020spotfast}. 

In this paper, we benchmark all of the above state-of-the-art methods on our dataset and present a detailed analysis of the results, which could provide directions for future research.

\begin{table*}[htb]
  \caption{LRWR benchmark}
  \label{tab:benchmarking}
  \centering
  \begin{tabular}{lcc}
    \toprule
    \multicolumn{1}{c}{\textbf{Network}} & 
    \multicolumn{1}{c}{\textbf{Initial weights}} &
    \multicolumn{1}{c}{\textbf{LRWR accuracy}} \\
    \midrule
    D3D\cite{DBLP:journals/corr/abs-1810-06990}   & LRW pretrained  & 54\%~~~             \\
    Feng20 \cite{feng2020learn}   & LRW pretrained  & 52\%~~~       \\
    Xiao20 \cite{xiao2020deformation}    & LRW pretrained  & 52\%~~~       \\
    Martinez20 \cite{martinez2020lipreading}    & LRW pretrained  & 51\%~~~       \\
    D3D\cite{DBLP:journals/corr/abs-1810-06990}    & none & 49\%~~~    \\
    CSN50 \cite{DBLP:journals/corr/abs-1904-02811} + GRU-backend + (2+1)D CNN & none  & 49\%~~~       \\
    CSN50 \cite{DBLP:journals/corr/abs-1904-02811} + GRU-backend & none  & 49\%~~~       \\
    Martinez20 \cite{martinez2020lipreading}    & none  & 48\%~~~       \\
    Xiao20 \cite{xiao2020deformation}    & none  & 47\%~~~       \\
    CSN34 \cite{DBLP:journals/corr/abs-1904-02811} & none  & 42\%~~~       \\
    \bottomrule
  \end{tabular}
\end{table*}

\section{Dataset}

In this section, we describe the properties of the LRWR dataset: the pipeline of raw data collection, its' preprocessing logic, and the final statistics of the resulting dataset.

\subsection{Data collection}

To make our benchmark suitable for a large variety of real-world use cases, we made a few requirements during the data collection process:
\begin{itemize}
\item First of all, we were carefully looking at the dataset diversity of speakers' audio/video conditions – different rates of speech, facial expressions, and the face's visual appearance.
\item Secondly, we were interested in different lipreading conditions – a variety of camera angles and lighting.
\end{itemize}

After an exploratory search, we found that YouTube videos could meet these requirements. 
Downloaded videos were available in HD (1280 * 720) or full\_HD (1920 * 1080), 25 fps, with 192 Kbps sound quality, and a variety of Russian-speaking podcasters and bloggers, resulting in a dataset of videos with a large diversity of speakers in different environments. Moreover, we were able to capture a wide range of topics in this case: history, art, travel, world events, or just daily talks.

We also apply several extra preparations during the data collection pipeline.
Firstly, we filter the videos to select one with only one speaker present in the frame.
Then we prepare a vocabulary with 1500 most frequent words of Russian-language blogging. The vocabulary was based on the first 50 hours of video from the previous step, distilled into text using the speech-to-text toolkit "Kaldi" \cite{ravanelli2019pytorchkaldi}.
We did the preliminary selection of 1500 words, because in Russian a word can have many forms, depending on the number, case, and conjugation. Because of this, many words appear only once, and are not suitable for inclusion in the dataset.

After these stages, we get a list of videos suitable for parsing, including 135 speakers and about 350 hours and the list of 1500 most frequently used words.

\subsection{Raw data preprocessing}
The data preprocessing pipeline is shown in Fig \ref{fig:speech_production}.
We process the downloaded audio track using speech-to-text methods and find the frames with words from our vocabulary.

As a speech-to-text baseline, we use the Kaldi toolkit \cite{ravanelli2019pytorchkaldi}, which outputs text and the appropriate time frame in which this text was spoken.  
After that, we cut this fragment from the video using FFmpeg\footnote{https://www.ffmpeg.org/}. 
Based on exploratory analysis of the data subset, we did not find any errors in extracted word boundaries, so we didn't have to implement video synchronization to get more accurate time coordinates.

Having received a video, we find the speaker's face and highlight the face's landmarks with the FaceAlignment network \cite{bulat2017far}. Then we normalize the video using landmarks, making the line between lips corners horizontal. Finally, we crop the lip's region during the entire video with an algorithm inspired by the paper \cite{DBLP:journals/corr/abs-1810-06990} using the following equation: 
\begin{equation}
  w = min(3d_{mn}, max(1.5d_{mn}, 1.05x_r - 0.95x_l))
  \label{eq1}
\end{equation}
where $w$ is a crop width and height, $d_{mn}$ is a distance between nose and center of lips and $x_l$, $x_r$ -- x coordinate of lip corners. To filter the inaccuracies of the FaceAligment network, we employ a binary classifier based on ResNet-18 \cite{DBLP:journals/corr/HeZRS15}. 
Moreover, to filter blurry footage or unuseful camera angles, we utilize the IoU tracker \cite{inproceedings} to check the face's sharp movement in the frame and filter such cases out.

The resulting dataset was extremely unbalanced: some classes had more than 2,000 words, most of them – less than 200. Moreover, some classes of words differed in the pronunciation of one letter. For instance, the dataset contains words pronounced \textit{kotoriy} and \textit{kotoria} as different classes. Our initial experiments demonstrate that too many classes that differ in only one phoneme lead to a significant deterioration in quality. In such a case, we leave only one class not to disrupt the neural networks learning process. After such a strict selection, there were 235 classes with more than 500 words were selected.

In the final step, we balance classes, leaving 500 words in each class. The resulting dataset was used for further benchmarking of the networks.

\subsection{Dataset Statistics}

The LRWR contains 235 words, 135 speakers with over 117500 samples. Each class represents by 450 train samples and 50 test samples. 
The dataset has a broad coverage of different ages, genders, and pronunciation habits of the speakers.
Lighting conditions range from natural daylight to artificial.
The face rotation angle varies from 0\textdegree to 20\textdegree. 
By these factors, the dataset presents a challenging task for current lipreading methods. 
Moreover, for the speaker's privacy presented in the data, we make publicly available\footnote{The dataset is available on request to authors upon submitting a license agreement.} only cropped speakers' lip area as a final benchmark dataset as shown in Fig \ref{fig:speech_production}.

\begin{table*}[htb]
  \caption{Training tips analysis on \textit{LRWR}}
  \label{tab:experiments2}
  \centering
  \begin{tabular}{lcccccccc}
    \toprule
    \multicolumn{1}{c}{\textbf{Network}} & 
    \multicolumn{1}{c}{\textbf{LRW weights}} &
    \multicolumn{1}{c}{\textbf{Label Smoothing}} &
    \multicolumn{1}{c}{\textbf{CutOut}} &
    \multicolumn{1}{c}{\textbf{MixUp}} &
    \multicolumn{1}{c}{\textbf{CosineScheduler}} &
    \multicolumn{1}{c}{\textbf{upsampling}} &
    \multicolumn{1}{c}{\textbf{gru size}} &
    \multicolumn{1}{c}{\textbf{LRWR accuracy}} \\
    \midrule
    D3D\cite{DBLP:journals/corr/abs-1810-06990}   & \checkmark & \checkmark  & \texttimes & \checkmark & \checkmark & $450 \rightarrow 750$ & 2048 & 61\%~~~       \\
    D3D\cite{DBLP:journals/corr/abs-1810-06990}   & \checkmark & \checkmark  & \texttimes & \checkmark & \checkmark & $450 \rightarrow 750$ & 512 & 59\%~~~       \\
    D3D\cite{DBLP:journals/corr/abs-1810-06990}   & \checkmark & \checkmark  & \texttimes & \checkmark & \checkmark & \texttimes & 512 & 55\%~~~       \\
    D3D\cite{DBLP:journals/corr/abs-1810-06990}   & \checkmark & \checkmark  & (1,32) & \texttimes & \texttimes & \texttimes & 512 & 54\%~~~            \\    
    D3D\cite{DBLP:journals/corr/abs-1810-06990}   & \checkmark & \checkmark  & (32,8) & \texttimes & \texttimes & \texttimes & 512 & 54\%~~~            \\
    D3D\cite{DBLP:journals/corr/abs-1810-06990}   & \checkmark & \checkmark  & (8,8) & \texttimes & \texttimes & \texttimes & 512 &  54\%~~~            \\
    D3D\cite{DBLP:journals/corr/abs-1810-06990}   & \checkmark & \checkmark  & \texttimes & \texttimes & \texttimes & \texttimes & 512  &  54\%~~~            \\
    D3D\cite{DBLP:journals/corr/abs-1810-06990}   & \checkmark & \texttimes  & \texttimes & \texttimes & \texttimes &\texttimes & 512 &  54\%~~~            \\
    D3D\cite{DBLP:journals/corr/abs-1810-06990}   & \checkmark & \checkmark  & \texttimes & \checkmark & \checkmark & $450 \rightarrow 950$ & 512 & 52\%~~~       \\
    D3D\cite{DBLP:journals/corr/abs-1810-06990}   & \texttimes & \texttimes  & \texttimes & \texttimes &\texttimes & \texttimes & 512 &  49\%~~~             \\
    \bottomrule
  \end{tabular}
\end{table*}

\section{Experiments}

In this section, we evaluate popular lipreading networks on our dataset, perform the analysis of training tricks effectiveness and summarise our finding into a new architecture proposal that achieved state-of-the-art accuracy on LRWR and LRW.

\subsection{Experimental settings}

For the benchmarking, the final version of the LRWR dataset with balanced classes was taken.
Additionally, we randomly flip all the frames in the video horizontally and randomly crop area (88*88) through all video, which is common practice for lipreading \cite{DBLP:journals/corr/abs-1810-06990, xiao2020deformation, martinez2020lipreading}. 
Furthermore, we explore neural networks training from random and pretrain weights initialization to estimate finetuning capabilities.

\subsection{Methods}

To provide a comprehensive benchmark, we review recent papers on lipreading, most common baselines, and the best-performing approaches in this task. We highlight three of them:
\begin{itemize}
\item 3D+2D frontend with recurrent layers backend approach, represented by the article "Learn an effective lipreading model without pains" \cite{feng2020learn} which currently has the highest score on the LRW and the LRW-1000.
\item Modifications to the previous method with recurrent layers replaced by the temporal convolution layers \cite{martinez2020lipreading}, Using the second branch, where 3D convolutions are replaced by the Deformation-Flow-Network \cite{xiao2020deformation}.
\item 3D-only frontend architecture, presented by the D3D \cite{DBLP:journals/corr/abs-1810-06990}.
\end{itemize}

Moreover, we adapted a method from the field of action recognition – the channel-separated convolutional networks \cite{DBLP:journals/corr/abs-1904-02811}. The adaptation includes usage of recurrent layers as a backend, following the idea presented in article \cite{DBLP:journals/corr/StafylakisT17}. 
As another approach for backend architecture, we replace 3D convolution with  (2+1)D-convolution \cite{DBLP:journals/corr/abs-1711-11248}. 
The final results of methods comparison are shown in the Table \ref{tab:benchmarking}.

\subsection{Analysis}

After selecting the most successful candidate - D3D, we investigated the impact of various regularization and augmentation techniques or learning process modification on the final quality.
We review the following techniques: Label Smoothing \cite{DBLP:journals/corr/SzegedyVISW15}, MixUp \cite{DBLP:journals/corr/abs-1710-09412}, 
CutOut \cite{DBLP:journals/corr/abs-1708-04552}. Moreover, we investigate the impact of the hidden size of the recurrent layers and Cosine scheduler usage \cite{DBLP:journals/corr/abs-1910-11605} during training. 
In addition, we examine upsampling. Since not all parsed data entered the LRWR, as mentioned in chapter 3.3, we add data that did not get into the LRWR due to class balancing. If there was not enough unused data, then the samples were copied.
The results of the experiments are presented in Table~\ref{tab:experiments2}. 

Experiments have shown that the combination of LabelSmoothing, MixUp, and CosineScheduler with increased RNN hidden size positively affects the resulting performance. 
We also found a general fact for LRWR: the quality of any network increases if we upsample each class up to 750 samples, while further upsampling leads to performance degradation. 
Moreover, pretrained weights initialization leads to better results in all our experiments that provide promising directions for lipreading models finetuning in future research.
Finally, while D3D demonstrates the greatest performance on the LRWR dataset (Table \ref{tab:benchmarking}), it has the lowest accuracy on the LRW (Table \ref{tab:example}), which could demonstrate the differences between the benchmarked languages 

\subsection{LRW improvement – AttentionLipreadingNet}

Using our findings from the LRWR benchmark, we would like to introduce a new architecture named AttentionLipreadingNet (ALN) and describe its training process, leading to the new state-of-the-art results on LRW dataset.

\textbf{AttentionLipreadingNet}.
Analyzing the differences between published (Table \ref{tab:example}) and benchmarked (Table \ref{tab:benchmarking}) results, we would like to propose a neural architecture that will work effectively on both datasets. Despite the fact that on LRWR dataset D3D\cite{DBLP:journals/corr/abs-1810-06990} shows the best results, this network demonstrate lower quality on the LRW compared to networks using a combination of 3D and 2D convolutions. Therefore, to test our developments on the LRW, we need to develop new network, and apply all the techniques that led to the accuracy improvement on the LRWR.
\par The new network was inspired by the paper \cite{feng2020learn}, which architecture shows the best results on the LRW and LRW-1000. The frontend of ALN consists of 3D convolutions followed by 2D ResNeSt-based convolutions with split-attention \cite{zhang2020resnest}, and a backend consisting of recurrent layers.

\textbf{Train Loop}. For each video submitted for training, a region of size 88 * 88 was randomly selected, and only a crop of this region was used for training. 
After that, we use horizontal flip augmentation with 50\% probability, followed by MixUp, LabelSmoothing, and DropBlock augmentations. We used Adam Optimizer with an initial learning rate of 1e-4, combined with CosineScheduler and a batch size of 32. LogSoftmax averaged over the length of the video was used as a loss. 

We trained the network on the LRW dataset for 60 epochs on the NVidia V100 GPU. As a result, we got an accuracy of 89.1\%, which is superior to the previous best result (Table~\ref{tab:example}).

\begin{table}[htb]
  \caption{Results on LRW and LRWR. All LRWR networks were pretrained on LRW}.
  \label{tab:example}
  \centering
  \begin{tabular}{ c  c  c}
    \toprule
    \multicolumn{1}{c}{\shortstack{\textbf{Method} }} & 
    \multicolumn{1}{c}{\shortstack{\textbf{LRW Accuracy}  }} &
    \multicolumn{1}{c}{\shortstack{\textbf{LRWR Accuracy} }}\\ 
    \midrule
    \textbf{ALN} (ours) &  \textbf{89.1\%} & \textbf{61\%}~~~              \\
    Feng20 \cite{feng2020learn} & 88.4\% & 52\%~~~              \\
    Martinez20 \cite{martinez2020lipreading} &  85.3\% & 51\%~~~ \\
    Xiao20 \cite{xiao2020deformation}&      84.1\% & 52\%~~~ \\
    D3D \cite{DBLP:journals/corr/abs-1810-06990} & 78.0\% & 54\%~~~             \\
    \bottomrule
  \end{tabular}
\end{table}

\par
After achieving this result on the LRW dataset, we train the ALN network on the LRWR using the same training settings and LRW pretrain weights, resulting in the 61.1\% accuracy.
We explain such a significant disparity (61\% and 89\% accuracy) in the final quality on LRW and LRWR due to the lip region quality difference: 112 (LRWR) versus 256 (LRW) (Table \ref{tab:datasets}), and more complex and varied shooting conditions. 
Another possible reason for the difference in the resulting quality may be the phonetic difference of the languages themselves and its impact on the lipreading methods' performance, which we see as another possible direction for future research.




\section{Conclusion}

In this paper, we introduce the first dataset for the lipreading benchmark in Russian language – LRWR, containing 235 classes with 117500 samples. We evaluate current lipreading methods and investigate the impact of different techniques on the final performance. As a result of our benchmark, we developed a network that accomplishes the best score on both LRWR and LRW datasets, improving the accuracy on the last one from 88.4\% to 89.1\% achieving new state-of-the-art results.

\section{Acknowledgements}

The authors are grateful to Konstantin Osminin and Pavel Kalaidin for useful discussions and valuable comments.

\bibliographystyle{IEEEtran}

\bibliography{template}

\end{document}